\documentclass{llncs}
\usepackage{amssymb}
\usepackage{mathrsfs, amsmath, float, multirow}
\usepackage[utf8x]{inputenc}
\usepackage[ruled, vlined, lined, linesnumbered]{algorithm2e}
\usepackage{graphicx, tabularx, subfigure, subfloat, wrapfig}
\usepackage{url}

\usepackage{color}
\usepackage{graphicx}

\begin{document}

\title{Learning Optimal Deep Projection of $^{18}$F-FDG PET Imaging for Early Differential Diagnosis of Parkinsonian Syndromes}

\author{Shubham Kumar$^{1,2,}$\thanks{Equal Contribution\protect\label{X};~$\dag$ corresponding to zuochuantao@fudan.edu.cn}, Abhijit Guha Roy$^{1,3,\star}$,  Ping Wu$^{4}$, Sailesh Conjeti$^{1,5}$, R. S. Anand$^{2}$, Jian Wang$^{6}$, Igor Yakushev$^{7}$, Stefan Förster$^{7}$, Markus Schwaiger$^{7}$, Sung-Cheng Huang$^{8}$, Axel Rominger$^{9}$, Chuantao Zuo$^{4,\dag}$ \and Kuangyu Shi$^{1,9}$}

\institute{$^1$Computer Aided Medical Procedures, Technical University of Munich, Germany \\
$^2$Indian Institute of Technology, Roorkee, India \\
$^3$Artificial Intelligence in Medical Imaging (AI-Med), KJP, LMU Munich, Germany\\
$^4$PET Center, Huashan Hospital, Fudan University, Shanghai, China \\
$^5$German Center for Neurodegenerative Diseases (DZNE), Bonn, Germany \\
$^6$Dept. Neurology, Huashan Hospital, Fudan University, Shanghai, China \\
$^7$Dept. Nuclear Medicine, Technical University of Munich, Germany \\
$^8$Dept. Molecular and Medical Pharmacology, University of California, LA, USA \\
$^9$Dept. Nuclear Medicine, University of Bern, Switzerland} 

\maketitle

\begin{abstract}
Several diseases of parkinsonian syndromes present similar symptoms at early stage and  no objective widely used diagnostic methods have been approved until now. Positron emission tomography (PET) with $^{18}$F-FDG was shown to be able to assess early neuronal dysfunction of synucleinopathies and tauopathies. Tensor factorization (TF) based approaches have been applied to identify characteristic metabolic patterns for differential diagnosis. However, these conventional dimension-reduction strategies assume linear or multi-linear relationships inside data, and are therefore insufficient to distinguish nonlinear metabolic differences between various parkinsonian syndromes. In this paper, we propose a Deep Projection Neural Network (DPNN) to identify characteristic metabolic pattern for early differential diagnosis of parkinsonian syndromes. We draw our inspiration from the existing TF methods. The network consists of a (i) compression part: which uses a deep network to learn optimal 2D projections of 3D scans, and a (ii) classification part: which maps the 2D projections to labels. The compression part can be pre-trained using surplus unlabelled datasets. Also, as the classification part operates on these 2D projections, it can be trained end-to-end  effectively with limited labelled data, in contrast to 3D approaches. We show that DPNN is more effective in comparison to existing state-of-the-art and plausible baselines.

\end{abstract}
\newcommand\T{\rule{0pt}{1.0ex}}
\newcommand\B{\rule[-1.2ex]{0pt}{0pt}}

\section{Introduction}

Approximately 7 to 10 million people worldwide are suffering from Parkinson’s disease (PD). On the other hand, very similar clinical signs can appear in patients with atypical parkinsonian syndromes, such as multiple system atrophy (MSA) and progressive supranuclear palsy (PSP) and these conditions account for approximately 25-30\% of all cases of parkinsonian syndromes~\cite{Hughes2001}. Diagnosis of parkinsonian patients  based on longitudinal clinical follow up remains problematic with a large number of misdiagnoses in early stage~\cite{Fahn2004}. Thus, early differential diagnosis is essential for determining adequate treatment strategies and for achieving the best possible outcome for these patients~\cite{LiR2017}. 

Positron emission tomography (PET) captures neuronal dysfunction of PD using specific \emph{in-vivo} biomarkers~\cite{Gao2012,Xu2014,BagciU2013,Jiao2014,Bi2014} and has been shown to be more advantageous in early diagnosis, far before structural damages to the brain tissue occurs~\cite{Jiao2014,Zhang2012,Zhou2012,LuS2017}. Automated approaches such as Principal component analysis (PCA) has been successfully applied on $^{18}$F-FDG PET to extract PD-related pattern (PDRP), MSA-related pattern (MSARP), and PSP-related pattern (PSPRP)~\cite{Eidelberg2009,Tang2010}. These patterns have been found as effective surrogates to discriminate between classical PD, atypical parkinsonian syndromes and healthy control subjects~\cite{Tang2010}. To account for heterogeneous physiology and enable individual pattern visualization, a tensor-factorization based method was developed by projecting the 3D data into 2D planes containing the discriminative information~\cite{LiR2017}. However, these conventional dimension-reduction based methods assume linear or multi-linear relationship inside data. In contrast, different subtypes of parkinsonian syndromes, caused by different protein aggregation ($\alpha$-synuclein or Tau),  show a non-linear relationship to the anatomical changes. Thus difference of metabolic patterns between PD, MSA and PSP can be nonlinear due to these diverse pathological manifestations and heterogeneous propagation among complex brain connectomes. Therefore, either PCA or tensor factorization is insufficient to identify nonlinear metabolic differences of various parkinsonian syndromes, and is susceptible to providing sub-optimal solutions.

Deep learning based approaches have recently been shown to be very effective in discovering non-linear characteristic patterns within data in an end-to-end fashion~\cite{LeCun2015,Ithapu2015}. 
It has been shown to surpass human performance in different complicated tasks, like image classification. 
It has also gained a lot of popularity in the bio-medical community \cite{LITJENS201760} for computerized diagnosis on medical imaging, such as differential diagnosis~\cite{Ithapu2015,ZhouL2011,Suk2015}. Inspired by these recent successes, we use a deep learning based architecture for early diagnosis of parkinsonism.

One of the major challenge associated with this task is that our input data is 3D in nature, with limited amount of labelled training samples. Standard approaches of going for 3D based CNN models (very high number of learnable parameters) are prone to overfitting when trained on limited samples. To circumvent this issue, we draw inspiration from the existing approaches which uses Tensor Factorization (TF) to project the 3D scans to 2D, and use them for diagnosis. Towards this end, we propose a deep projection neural network (DPNN), which has two parts, (i) Compression Part and (ii) Classification Part. The Compression Part basically mimics TF projection from 3D to 2D. This part can be pre-trained on a large amount of unlabelled dataset, which is easily available. This pre-trained model is added to the 2D-CNN based Classification part (lower model complexity), which is trained end-to-end with limited annotated data. Although in this paper, we present its application for PET scans, the concept is fairly generic and can be easily extended to any 3D data.

\section{Materials \&  Methods}

\subsection{Data Preparation \& Preprocessing}
A cohort of 257 patients (Dataset-1) with clinically suspected parkinsonian features were included in this study. The patients were referred for $^{18}$F-FDG PET imaging and then assessed by blinded movement disorders specialists for more than 2 years. Finally 136 of them were diagnosed with PD, 91 with MSA and 30 with PSP. All the 3D PET volumes were preprocessed using intensity-normalized by global mean and spatially normalized to Montreal Neurological Institute (MNI) space using SPM8\footnote{Statistical Parametric Mapping, \url{http://www.fil.ion.ucl.ac.uk/spm/software/spm8/}, 2009} according to a standard PET processing procedure~\cite{LiR2017}.  For optimizing deep networks, the limited availability of PET images of patients at early stage of parkinsonism could be a bottleneck. Therefore, a database of 1077 subjects (Dataset-2) with 41 various non-parkinsonian neurological diseases with brain FDG PET images is further included to enhance the data pool.

\subsection{DPNN Architecture}
\begin{figure}[tb]
\centering
\includegraphics[width=1.0\textwidth]{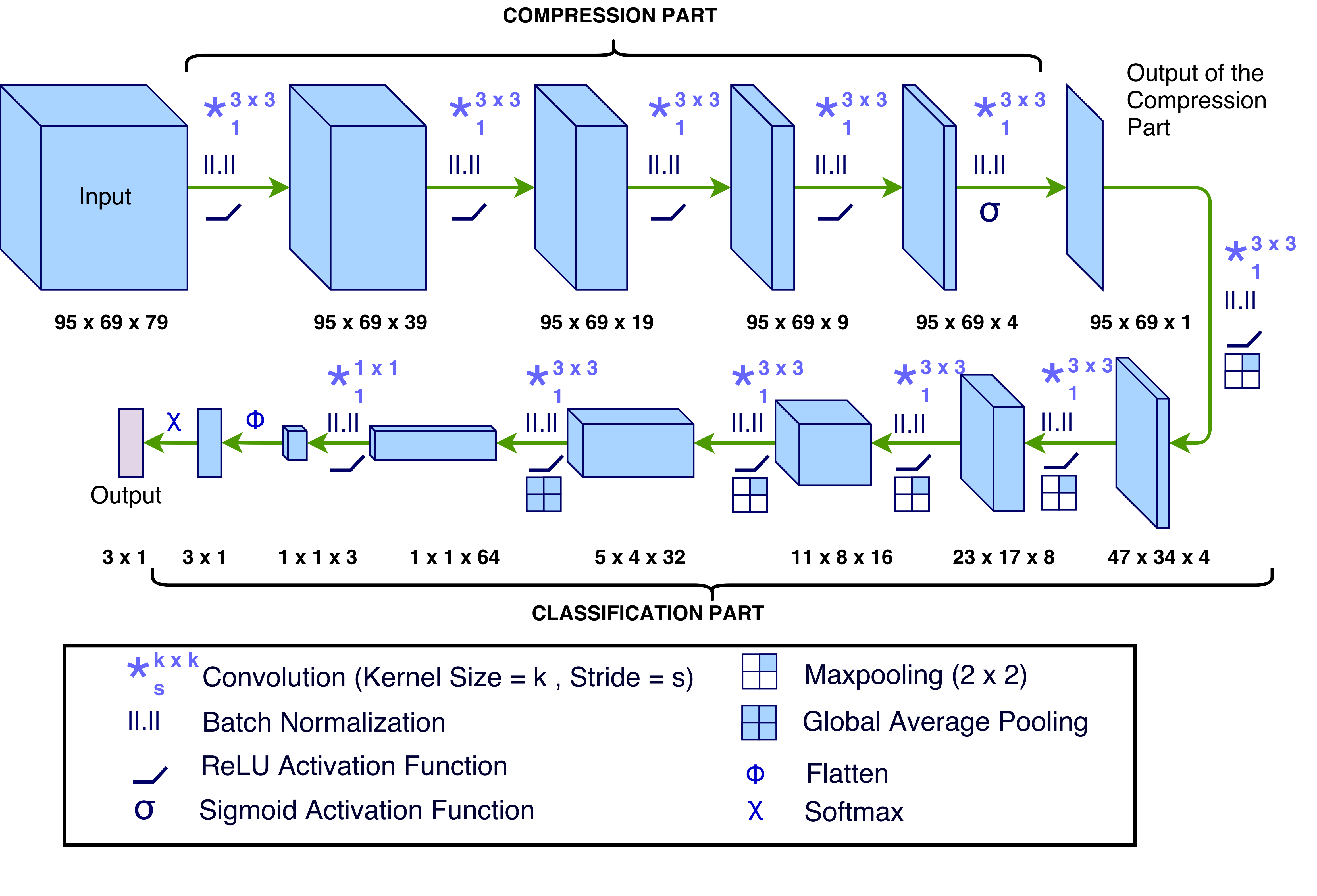}
\caption{Illustration of the overall model architecture of deep projection neural network (DPNN). All the architectural details regarding the network are shown here.}
\label{fig:dpnn}
\vspace{-5mm}
\end{figure}

We draw our inspiration from prior work which estimated tensor factorized projection of 3D PET scans and processed them for classification task. In this regard, we formulated to solve the problem in two parts: (i) Learn a separate network to mimic the tensor factorization from 3D data, \textit{i.e.} learning to compress the data (Compression Part), and (ii) Learn a 2D CNN model to map the compressed input to one of the classes (Classification Part). A detailed description of both the parts are provided below with the architectural design in Fig.~\ref{fig:dpnn}.

\noindent
\textbf{Compression Part:}
Given a 3D PET scan $I_P \in \mathbb{R}^{H \times W \times D}$, here we estimate a function $f_p(\cdot)$ which compresses the data to a 2D projection map $P_t$, so that $f_p:I_P\rightarrow P_t$, where $P_t \in \mathbb{R}^{H \times W}$. This non-linear function $f_p(\cdot)$ is approximated by a series of blocks consisting of a $3 \times 3$ convolutional layer, batch normalization and a ReLU activation function. A set of 5 such blocks are stacked together, which compresses $I_P$ sequentially to $P_t$. The final block uses a sigmoidal non-linearity instead of ReLU to rescale the activations between $[0,1]$.

\noindent
\textbf{Classification Part:}
This part takes $P_t$, the compressed projection map as input. It learns a mapping $f_c(\cdot)$, which maps $P_t$ to the one of the class labels $y$. The first 5 blocks consist of a $3 \times 3$ convolutional layer, batch norm, ReLU activation and a max pooling layer, reducing the spatial dimensions by a factor of 2 at every step. The final block consists of a global average pooling instead of max pooling, squeezing the feature map along spatial dimensions.This is followed by a $1 \times 1$ convolutional layer, softmax layer to project the learnt features to the label probalility space $\mathbb{R}^3$, from where $y$ is estimated as the class with highest probability. More details regarding the size of intermediate feature maps and stride are indicated in Fig.~\ref{fig:dpnn}.

\subsection{Training Procedure}
To tackle the issue of learning such a highly complex model with limited training data, we propose to address the training procedure in two stages: (i) We leverage unlabelled PET data corpus to pre-train the Compression Part, (ii) limited labelled data is used to learn the weights of the Classification Part, with the Compression Part initialized to the pre-trained weights.

\noindent
\textbf{Pretraining: }
In this part, we use the unlabelled Dataset-2 $\{I_i\}$ for pre-training. We compute the tensor factorized 2D maps of all the volumes as $\{ \mathcal{G}_i \}$. In this stage, we train the compression part $f_p(\cdot)$, using this dataset, with the goal of mimicking $\{ \mathcal{G}_i \}$ as the output of the network. We hypothesize that this provides a strong initialization to the network for the classification stage. The network is learnt by jointly optimizing a combination of Mean Square Error (MSE) and Structural Similarity Index (SSIM) between the target and prediction, defined as,

\begin{equation}
\label{eqn:cost1}
\mathcal{L}=\underbrace{
\frac{1}{2N_p}\sum_{\mathbf{r}}{(\mathcal{P}(\mathbf{r})-\mathcal{G}(\mathbf{r}))^2}}_{\mathrm{MSE}
}- \underbrace{
\frac{1}{N_w}\sum_{\mathbf{w}}{\rm SSIM}({\mathbf{w}_p},{\mathbf{w}_g}),
}_{\mathrm{SSIM \ Index} 
}
\end{equation}

\begin{equation}
{\rm SSIM}({\mathbf{w}_p},{\mathbf{w}_g})={\frac{(2\mu_{p}\mu_{g}+C_{1})(2\sigma_{pg}+C_{2})}{(\mu_{p}^{2}+\mu_{g}^{2}+C_{1})(\sigma_{p}^{2}+\sigma_{g}^{2}+C_{2})}},
\end{equation}

\noindent
where, $\mathcal{P}$, $\mathcal{G}$, $\mathbf{r}$ and $N_r$ are the predicted map, target projection map, pixel-position, and the total number of pixels respectively. ${\mathbf{w}_p}$ and ${\mathbf{w}_g}$ represent a local $6\times6$ window in $\mathcal{P}$ and $\mathcal{G}$, and ${N_w}$ is the total number of such windows. SSIM is calculated on all the ${N_w}$ windows and their average value is used in the cost function.  $\mu_{p}$, $\sigma_{p}^2$, and $\sigma_{pg}$ are the mean of ${\mathbf{w}_p}$, the variance of ${\mathbf{w}_p}$, and the covariance of ${\mathbf{w}_p}$ and ${\mathbf{w}_g}$, respectively. $C_1$ and $C_2$ are set to $\sim 10^{-4}$ and  $\sim 9\times10^{-4}$, respectively. We use SSIM based loss function to preserve the quality of the predicted map similar to actual Tensor Factorized map. The weights of the convolutional kernels are initialized using Xavier initialization and Adam optimizer with  a learning rate of $10^{-4}$ is used for the weight updates. The $\beta_1$, $\beta_2$ and $\epsilon$ parameters of the optimizer are set to $0.9$, $0.999$, and $10^{-8}$, respectively. The training is continued until the validation-cost saturates. 

\noindent
\textbf{Fine-Tuning: }
In this stage, the pre-trained compression network is combined with the classification part, and the weights of the classification part are initialized using Xavier initialization. The whole network is trained in an end-to-end fashion, minimizing 3-class Cross-Entropy loss function using Adam optimizer with  $\beta_1$, $\beta_2$ and $\epsilon$ set to $0.9$, $0.999$, and $10^{-8}$, respectively. The learning rate used for the classification part is $10^{-4}$, while for the compression part it is kept $10^{-5}$. The learning rate of the compression part is kept one order low to prevent high perturbation in those layer. The weights are regularized with a decay constant of $10^{-5}$, preventing over-fitting. A mini-batch of $10$ PET scans are used. The training is continued until the convergence of the validation loss.

\section{Experiments and Results}%
\label{sec:expRes}%

\subsubsection{Experiments:}
We evaluate our proposed DPNN model by a 5-fold cross-validation experiment on Dataset-1. An equal distribution of each of samples from the three classes were ensured in each of the folds. For evaluation, we used the standard metrics, (i) True Positive Rate (TPR), (ii) True Negative Rate (TNR), (iii) Positive Predictive Value (PPV), and (iv) Negative Predictive Value (NPV), consistent with \cite{LiR2017}.

\noindent
\textbf{Baselines: }
We compare our proposed method against state-of-the-art method which uses Tensor Factorization (TF), followed by SVM for classification~\cite{LiR2017}. Apart from this, we define two other  baselines to substantiate our claims:

\begin{enumerate}
\item \textbf{BL-1: }DPNN, with pre-training using only MSE, to observe the effect of including SSIM in the cost function.

\item \textbf{BL-2: }DPNN, without pre-training, trained end-to-end, to observe the effect of pre-training.
\end{enumerate}
For all the experiments we used five-folded cross-validation for evaluation. All the networks were trained on NVIDIA Titan-Xp GPU with 12GB RAM.

\subsubsection{Results:}

\begin{figure}
	\centering
\includegraphics[width=0.9\linewidth]{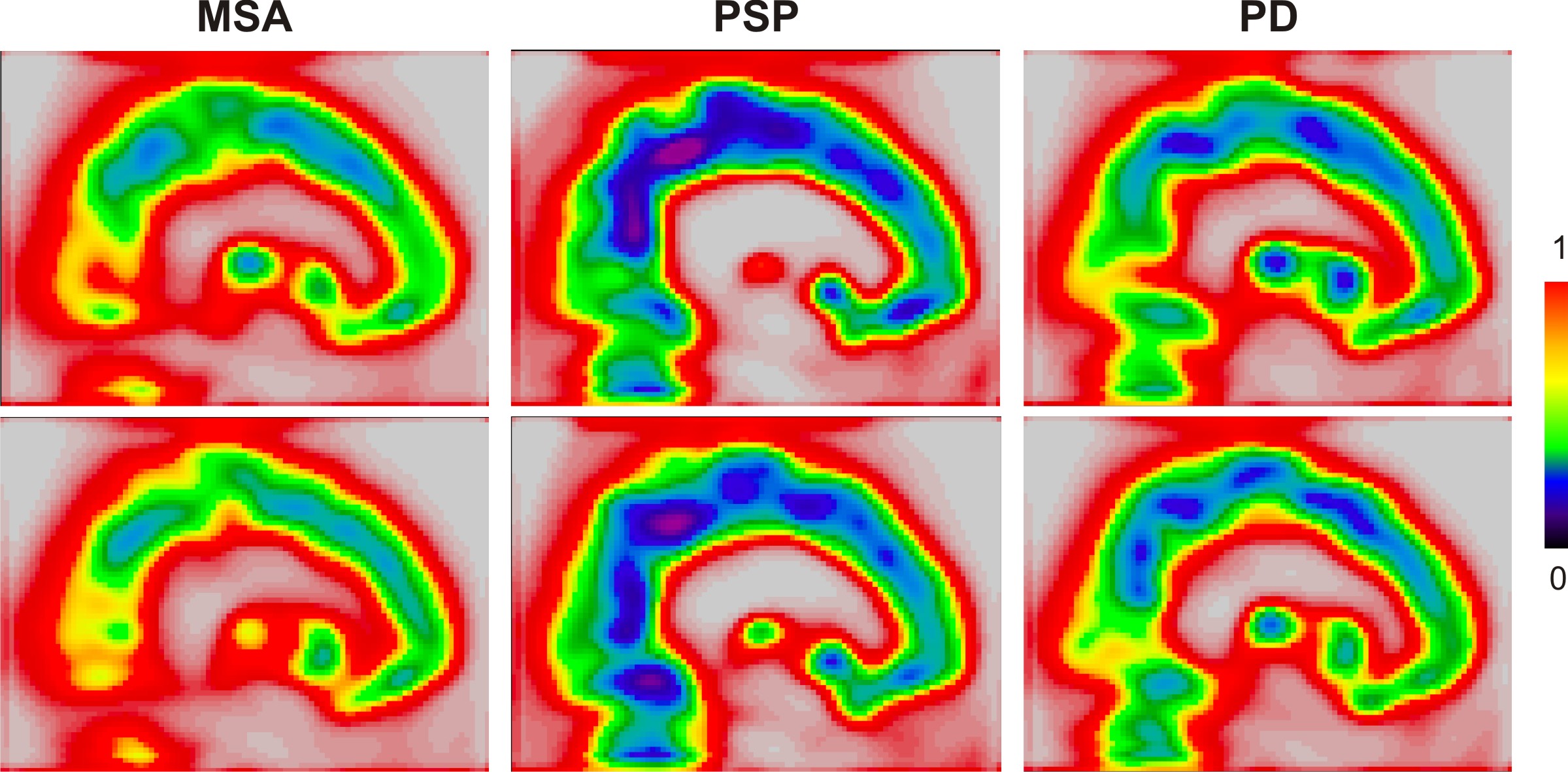}	

\caption{Projection of 3D PET Volumes generated by the Compression Part of the fine-tuned DPNN. We can visually observe the distinct patterns exhibited by the three sub-types MSA, PSP and PD, which not only aids clinicians for inference, but also aids the Classification part in automated decision making. }
\label{fig:projections}
\end{figure}

Tab.~\ref{tab:results} reports the results of our proposed model (DPNN), with defined baselines and state-of-the-art method, in terms of the mentioned evaluation metrics. Comparing with state-of-the-art~\cite{LiR2017}, DPNN outperforms it in most of the evaluation scores (8 out of 12). Comparing with \textbf{BL-1}, DPNN outperforms it. This substantiates our previous hypothesis that MSE+SSIM based pre-training is more effective in providing stronger initialization than MSE alone, which fails to capture the quality based features in the compression stage. It can be attributed to the fact that SSIM applies stricter constraint on similarity which forces the network to learn better representations. Also, comparing to \textbf{BL-2}, we prove our previous claim that pre-training is necessary when training such a complicated model with limited annotated data. It has improved the specificities of MSA by 0.79\%, PSP by 17.97\% and PD by 14.37\%, which can play a critical role for differential diagnosis. It is worth noting that DPNN shows consistent good performance across all the metrics for the PD class which has the highest number of samples (\textit{viz.} 136). While all the models show greatest inconsistency in the scores for PSP class, which has just $30$ representative samples in the dataset. This is indicative of the fact that given enough data the performance of DPNN can be increased to an ideal level.\newline\newline
Next, we take a closer look at the learnt Projection Maps in Fig.~\ref{fig:projections}, which shows example pattern images of MSA, PSP and PD. Patterns similar to tensor-factorization have been observed in the DPNN Projection results, for example, visible cerebellum and striatum activities in PD, vanishing cerebellum and striatum activities in MSA and decreasing striatum activity and visible cerebellum activity in PSP~\cite{LiR2017}. This confirms that DPNN is capable of extracting physiologically meaningful patterns, and use it for final decision making.

\begin{table}[htp]
\centering
\caption{Classification results of our proposed DPNN, in comparison to comparative methods and Baselines.}
\label{tab:results}
\resizebox{\textwidth}{!}{%
\begin{tabular}{|p{3.7cm}|c|c|c|c|c|c|c|c|c|c|c|c|}
\hline
\multicolumn{1}{|c|}{\multirow{3}{*}{Model}} &
\multicolumn{12}{c|}{Metrics} \\ \cline{2-13}
 \multicolumn{1}{|c|}{} & \multicolumn{4}{c|}{MSA} & \multicolumn{4}{c|}{PSP} & \multicolumn{4}{c|}{PD} \\ \cline{2-13}
\multicolumn{1}{|c|}{} & TPR & TNR & PPV & NPV & TPR & TNR & PPV & NPV & TPR & TNR & PPV & NPV \\ \hline

\textbf{DPNN} & 84.56 & $\mathbf{94.58}$ & 89.63 & $\mathbf{91.83}$ & 90.00 & $\mathbf{96.93}$ & 79.29 & $\mathbf{98.67}$ & $\mathbf{94.87}$ & $\mathbf{93.33}$ & $\mathbf{94.28}$ & $\mathbf{94.24}$\\ \hline

\textbf{BL-1} & 76.78 & 93.98 & 89.83 & 88.52 & 86.67 & 96.04 & 76.9 & 98.24 & 92.65 & 86.67 & 89.44 & 91.57\\ \hline

\textbf{BL-2} & 75.67 & 93.40 & 87.90 & 87.97 & 80.00 & 96.48 & 79.25 & 97.40 & 91.96 & 83.33 & 86.87 & 90.46\\ \hline

\textbf{TF+SVM}~\cite{LiR2017} & 86.35 & 93.79 & 92.85 & 88.86 & 97.87 & 78.96 & 97.30 & 85.07 & 92.44 & 78.96 & 89.14 & 85.73\\ \hline

\end{tabular}
}
\end{table}

\section{Conclusion}
We developed a deep learning method to extract characteristic metabolic pattern for differential diagnosis of parkinsonian syndrome. In contrast to linear or multi-linear data-reduction methods of the state-of-the-art, the proposed DPNN, processes 3D-data using 2D-convolutions, can explore the non-linear metabolic differences between the subtypes. Furthermore, we introduced a training procedure based on the optimization of SSIM along with MSE which leverages tensor-factorized maps of inputs, from a domain similar to the task-input domain, to overcome the difficulties posed by a small dataset. With limited amount of data, the novel method has already achieved superior accuracy compared to the state-of-the-art. The advanced pre-training strategies play a critical role in the success of this novel method, which prevent the abort of cutting-edge developments before approaching to a large data-bank. The positive performance of deep learning in this study encourages a multi-center study, which is actively in preparation. Although the DPNN patterns extracted in this proof-of-concept study look similar to the previous tensor factorization, an extensive inspection by clinicians may discover the characteristic difference matching to improve accuracy.  With the increase of data access, the ability of the deep learning methods to discover new discriminative features will be enhanced, which may provide the potential for a diagnosis at even earlier stage before motor impairment appears, i.e. at prodromal parkinsonian stage such as rapid eye movement (REM) sleep behavior disorder (RBD).  

\footnotesize
\bibliographystyle{unsrt}
\bibliography{MICCAI_TUM}

\end{document}